\documentclass[11pt]{article}

\usepackage[a4paper,margin=1in]{geometry}
\usepackage{fontspec}
\usepackage{amsmath,amssymb}
\usepackage{booktabs}
\usepackage{array}
\usepackage{graphicx}
\usepackage{xcolor}
\usepackage{microtype}
\usepackage[hidelinks]{hyperref}
\usepackage{enumitem}
\usepackage{caption}
\newfontfamily\thaifont[
  Path=./,
  UprightFont=THSarabunNew.ttf,
  BoldFont=THSarabunNew-Bold.ttf,
  ItalicFont=THSarabunNew-Italic.ttf,
  BoldItalicFont=THSarabunNew-BoldItalic.ttf,
]{THSarabunNew}
\newcommand{\tha}[1]{{\thaifont #1}}

\hypersetup{
  colorlinks=true, linkcolor=black, citecolor=black, urlcolor=blue!60!black,
  pdftitle={TSWAP: A Multilingual Retrieval-Augmented Thai Wellness Advisor},
}
\title{\vspace{-2em}\bfseries TSWAP: A Multilingual Retrieval-Augmented\\ Thai Wellness Advisor}

\author{
  Pornthep Ukosaramig\thanks{Corresponding author. Project lead, contract C05F680144.}\\
  \small Digital Touch Point Co., Ltd.\\
  \small \texttt{aun\_@hotmail.com}
  \and
  Kobkrit Viriyayudhakorn\\
  \small iApp Technology Co., Ltd.\\
  \small \texttt{kobkrit@iapp.co.th}
}
\date{\today}

\begin{document}
\maketitle

\begin{abstract}
\noindent
We present TSWAP, a deployed eight-language conversational wellness advisor grounded, via
retrieval-augmented generation, in a verified knowledge base of Thai traditional medicine and
certified wellness providers. An unmodified open-weight LLM (Qwen3.6-35B-A3B on vLLM) is grounded
on a $\sim$30.6K-chunk Thai index by a hybrid dense--sparse retriever with cross-encoder
reranking; a first-turn query classifier \emph{forces} tool-based retrieval for entity lookups; a
rule-based safety layer enforces medical scope and Thai emergency routing; and all eight languages
are served zero-shot with translate-then-retrieve. We release the first Thai
traditional-medicine/wellness retrieval benchmark (50 questions with gold document IDs; Recall@5
$=0.88$), production QA logs (91.1\% test--retest pass over 259 cases), and a 71-question frontier
no-retrieval probe showing what each grounding pillar contributes: without the safety prompt the
backend model family produced a full drug-dosing schedule and complied with out-of-scope requests,
and without the knowledge base it produced zero verifiable provider recommendations. We further
report two transferable deployment findings: English-calibrated 4-bit AWQ quantization corrupts
Thai tone marks, and forced-retrieval routing is necessary for reliable grounding.

\vspace{0.6em}
\end{abstract}

\section{Introduction}
Wellness tourism is among the fastest-growing segments of the global economy; Thailand's wellness
economy was valued at US\$40.5 billion ($\approx$1.4 trillion THB) in 2023, with wellness-tourism
spending of US\$12.3 billion \cite{gwi2025}. Yet information about Thai wellness and beauty
providers---services, packages, prices, certification status, and the traditional-medicine and
herbal knowledge that differentiates Thai wellness---remains fragmented across channels, with no
trustworthy, verifiable, multilingual central source. Foreign visitors in particular face a
language barrier and uncertainty about provider standards.

The Thai Smart Wellness Advisor Platform (TSWAP) was built to address this gap: a web, mobile, and
chatbot platform that consolidates verified provider data and Thai traditional-medicine knowledge
and exposes it through a multilingual conversational advisor. This paper focuses on the
conversational AI engine.

From an NLP standpoint, TSWAP occupies an empty niche. A survey of Thai and Southeast-Asian NLP
(Section~\ref{sec:related}) finds a crowded field of general-purpose Thai LLMs
\cite{typhoon,typhoon2,openthaigpt15,sealion,sailor,chinda} and a small body of Thai
\emph{clinical} NLP \cite{eir}, but \emph{no public NLP benchmark or dataset for Thai
traditional or herbal medicine} (\tha{แพทย์แผนไทย} / \tha{สมุนไพรไทย}), and only one prior Thai
traditional-medicine language model---AppHerb \cite{appherb}, a Gemma-2 fine-tune that
\emph{generates} treatment and recipe text from two textbooks, with no deployment, no
multilingual support, no retrieval grounding, and no reusable benchmark. The mature analogue is
Traditional Chinese Medicine, which has a developed benchmarking subfield
\cite{mtcmb,tcmladder}; Thai has essentially no equivalent.

\paragraph{Contributions.}
\begin{enumerate}[nosep,leftmargin=1.4em]
  \item \textbf{The first public Thai traditional-medicine/wellness retrieval benchmark} (50
  questions, gold document IDs, Recall@5 $=0.88$), released together with production QA logs and
  frontier-probe logs (Section~\ref{sec:eval}).
  \item \textbf{A grounding-first RAG recipe} whose central element---a query classifier that
  \emph{forces} retrieval for entity lookups---we show is necessary to prevent silent
  hallucination, with transferable serving findings: English-calibrated AWQ quantization corrupts
  Thai output, and hidden-reasoning budgets silently truncate answers
  (Sections~\ref{sec:rag},~\ref{sec:eval}).
  \item \textbf{A production evaluation with a Thai-localized safety probe} quantifying what the
  safety prompt and the knowledge base each contribute (Section~\ref{sec:probe}).
\end{enumerate}
TSWAP's LLM is deliberately \emph{not} fine-tuned and its safety layer is prompt-and-routing
only; we treat these as design choices with measurable consequences.

\section{Related Work}\label{sec:related}
\paragraph{Thai and SEA LLMs.} Open Thai/SEA foundation models include Typhoon
\cite{typhoon,typhoon2}, OpenThaiGPT \cite{openthaigpt15}, SEA-LION \cite{sealion}, Sailor
\cite{sailor}, and Chinda \cite{chinda}. These are general-purpose; none
targets traditional or herbal medicine. TSWAP uses an unmodified multilingual base model
(Qwen3.6-35B-A3B) and invests in grounding rather than pretraining.

\paragraph{Thai medical and traditional-medicine NLP.} Eir \cite{eir} is a Thai clinical LLM
evaluated on clinical tasks; evaluations of frontier models on the Thai national medical
licensing examination exist \cite{thainle2024}. General Thai benchmarks
(ThaiExam~\cite{typhoon}, M3Exam~\cite{m3exam}, WangchanThaiInstruct~\cite{wangchanx}) are
exam- or general-domain. The only Thai traditional-medicine LM is AppHerb \cite{appherb}. None
evaluate open-ended, culturally specific herbal/wellness advisory quality, and none is deployed
and multilingual.

\paragraph{Traditional-medicine benchmarks (analogue).} Traditional Chinese Medicine has multiple
benchmarks---MTCMB \cite{mtcmb}, TCM-Ladder \cite{tcmladder}---which we take as templates for a
future Thai herbal benchmark.

\paragraph{Multilingual health chatbots and evaluation.} Cross-lingual health-QA studies show
substantial non-English quality degradation \cite{xlingeval}; multilingual health assistants have
been studied in other low-resource settings \cite{healthpariksha}. For RAG we adopt
faithfulness/relevance/context metrics \cite{ragas} and medical-RAG evaluation practice
\cite{medrag}; for open-ended health-answer quality and safety we follow rubric-based grading
and safe/problematic/unsafe labeling \cite{healthbench,unsafe2025}. These inform our proposed
protocol (Section~\ref{sec:proposed}).

\section{The TSWAP Engine}\label{sec:system}
TSWAP is a production platform, publicly launched 28 June 2026 and generally available since
August 2026 on the web and as \emph{Wellness Chatbot} on Google Play and the App
Store.\footnote{\url{https://chat.thaismartwellnessai.com};
\url{https://play.google.com/store/apps/details?id=wellness.iapp.co.th};
\url{https://apps.apple.com/th/app/wellness-chatbot/id6756110840}} This paper concerns only its
\emph{AI engine}, which receives each user message with a data-minimized profile (wellness-relevant
fields only; no direct identifiers, for PDPA compliance) and owns the retrieval tools
(Section~\ref{sec:rag}), the multilingual behavior (Section~\ref{sec:multilingual}), and the
safety prompt (Section~\ref{sec:safety}).

\subsection{Knowledge base}\label{sec:kb}
The retrieval index is a single Milvus (v2.6.4) vector store holding eight collections
(Table~\ref{tab:collections}): provider and venue directories (restaurants, hotels, attractions,
wellness facilities, providers), curated packages and catalogs, and---central to the paper's
novelty---a \textbf{Thai herbal-medicine collection} (\texttt{thai\_herbs}). As of 2026-07-15 the
live index holds $\sim$30{,}625 chunks; venue collections are one chunk per entity, while
\texttt{thai\_herbs} and \texttt{attractions} are multi-chunk, so the number of distinct herb
monographs is smaller than the collection's chunk count.

\begin{table}[t]
\centering\small
\caption{Live retrieval index (chunks per collection, 2026-07-15).}
\label{tab:collections}
\begin{tabular}{lr@{\qquad}lr}
\toprule
Collection & Chunks & Collection & Chunks\\
\midrule
restaurants   & 11{,}212 & \texttt{thai\_herbs} & 870\\
hotels        & 8{,}704  & providers            & 484\\
attractions   & 8{,}169  & staygold\_catalogs   & 18\\
packages      & 1{,}153  & wellness\_facilities  & 15\\
\midrule
\multicolumn{4}{r}{\textbf{Total $\approx$ 30{,}625 chunks / 8 collections}}\\
\bottomrule
\end{tabular}
\end{table}

Each herb entry is a structured monograph (Table~\ref{tab:herbschema}). Two field groups matter
most for safe advisory behavior: \texttt{contraindications} (vulnerable-group cautions) and the
two reference fields (surfaced as citations, Section~\ref{sec:rag}). Ingestion is deterministic,
with per-collection SHA-256 change detection, guarded blue--green index rebuilds, and geocoding
enrichment for venues.

\begin{table}[t]
\centering\small
\caption{Schema of a herb monograph in \texttt{thai\_herbs}.}
\label{tab:herbschema}
\begin{tabular}{@{}ll@{}}
\toprule
Group & Fields\\
\midrule
Identity & \texttt{thai\_name}, \texttt{english\_name}, \texttt{scientific\_name}, \texttt{category}\\
Content  & \texttt{properties}, \texttt{benefits}, \texttt{usage\_instructions},\\
         & \texttt{botanical\_characteristics}, \texttt{pharmacology}\\
Safety   & \texttt{contraindications}\\
Evidence & \texttt{scientific\_references}, \texttt{research\_references}\\
Metadata & \texttt{origin\_region}, \texttt{usage\_type}, images\\
\bottomrule
\end{tabular}
\end{table}

\paragraph{Provenance.} The herbal collection is compiled from reference materials of the
Department of Thai Traditional and Alternative Medicine (DTAM,
\tha{กรมการแพทย์แผนไทยและการแพทย์ทางเลือก}), Ministry of Public Health---Thailand's authoritative
body for Thai traditional medicine---and each entry carries structured
\texttt{scientific\_references}/\texttt{research\_references} fields pointing back to source
literature. Chunking is character-based with a maximum of 512 tokens per chunk and 50-token
overlap.

\subsection{Retrieval pipeline}\label{sec:rag}

\paragraph{Base model and serving.} The generator is \textbf{Qwen3.6-35B-A3B}, a
mixture-of-experts model ($\sim$35B total / $\sim$3B active parameters), \emph{not fine-tuned},
served self-hosted with vLLM behind an OpenAI-compatible endpoint; ``thinking'' mode is disabled
in production. The deployed checkpoint is a community 4-bit AWQ quantization; we observed Thai
token corruption attributable to English-calibrated AWQ (Section~\ref{sec:eval}) and are migrating
to the official FP8 checkpoint.

\paragraph{Hybrid retrieval.} Retrieval is hybrid dense--sparse. Dense embeddings use
\texttt{BAAI/bge-m3} (1024-d, multilingual); sparse retrieval uses Milvus's built-in BM25 over raw
text. Per query, dense ANN (cosine) and BM25 (\texttt{drop\_ratio\_search}~$=0.1$) results are
fused with Reciprocal Rank Fusion; 10 candidates are retrieved, the top 7 are reranked by a
cross-encoder (\texttt{Qwen/Qwen3-Reranker-8B}), reranking is skipped when the top similarity
$\geq 0.85$, and the final \texttt{top\_k} is 3. Beyond semantic search, the engine applies
structured filters---province, food type, and price ranges with schema normalization---and
\textbf{geospatial filtering}: a nullable WKT \texttt{GEOMETRY} field with an RTREE index supports
\texttt{ST\_DWITHIN} radius queries, and a \texttt{near} parameter resolves a place name to
coordinates server-side.

\paragraph{Grounding-first routing.} A first-turn LLM query classifier routes each message to one
of five intents: wellness lookup, general advice, emergency, out-of-scope, or small talk. For
wellness lookups the engine forces the
retrieval tool via a named \texttt{tool\_choice}, guaranteeing that any entity lookup (herb,
provider, spa, package, hotel, restaurant, place) calls \texttt{rag\_search} before the model
answers. This fixes a concrete failure mode: with \texttt{tool\_choice=auto}, the model silently
skipped retrieval on short noun-phrase queries and answered from parametric memory. The system
prompt further constrains the model to recommend only items present in retrieval results, to never
invent entities, and to say ``no match found'' rather than fabricate. Only pure general-lifestyle
advice with no lookup target is answered without retrieval; the system is thus
\emph{grounded-for-entities} rather than strictly closed-book. For the longevity advisory variant,
the prompt mandates a trailing ``Citation:'' block of 3--5 items drawn from the KB's
\texttt{scientific\_references}/\texttt{research\_references} fields.

\subsection{Multilingual strategy}\label{sec:multilingual}
TSWAP serves eight languages---Thai, English, Chinese, Japanese, Malay, Russian, Korean, and
Hindi---with \emph{one} multilingual model and a \emph{single} English system prompt; there are no
per-language templates. The model replies in the language of the user's most recent message.
Because the knowledge base is Thai, retrieval uses a \textbf{translate-then-retrieve} strategy: the
model translates the query to Thai before calling \texttt{rag\_search}, then answers in the user's
language (bge-m3's multilinguality provides additional cross-lingual tolerance). All eight
languages are served \emph{zero-shot}; per-language output quality therefore rests entirely on the
base model's pretraining, which motivates the per-language evaluation we propose for the secondary
languages (Malay, Russian, Korean) in Section~\ref{sec:proposed}.

\subsection{Safety layer}\label{sec:safety}
The engine's server-owned system prompt implements a rule-based safety layer with four elements:
\begin{itemize}[nosep,leftmargin=1.2em]
  \item \textbf{Scope.} Wellness topics only (Thai herbs, nutrition, lifestyle, sleep, exercise,
  stress, providers, packages, wellness travel); anything else is politely declined and redirected.
  \item \textbf{Medical safety.} ``You are NOT a doctor'': no diagnosis, no prescribing, no drug
  dosages; guidance is framed as general wellness; the model recommends consulting a qualified
  professional, explicitly for children, pregnant/breastfeeding women, the elderly, and people with
  chronic illness or on medication.
  \item \textbf{Emergencies.} Chest pain, breathing difficulty, severe bleeding, stroke signs, or
  fainting trigger an immediate instruction to call 1669; self-harm or suicidal ideation triggers an
  empathetic response with hotline 1323. Emergencies are answered immediately, \emph{bypassing
  retrieval} (enforced both by the prompt and by the query classifier's \texttt{emergency}
  category).
  \item \textbf{Additive grounding.} Safety rules are additive and never replace retrieval,
  preventing safety text from being used to dodge grounding.
\end{itemize}
Two server-side mechanisms complement the prompt: the query classifier (above), and a
post-generation \textbf{Thai quality guard} that detects corrupted Thai via combining-mark
heuristics (mark-to-consonant ratio threshold $0.28$ plus two hard signals) and regenerates once
with official instruct-mode sampling. We stress that this is a \emph{quality} filter, not a safety
filter: there are \emph{no} post-generation safety classifiers, regex filters, or blocklists.
Herb--drug interaction risk is handled by deferral plus the KB's \texttt{contraindications} field
rather than a dedicated interaction checker. We report this plainly as a design choice and
limitation: the harm categories addressed are exactly those encoded in the prompt and routing
above (medical scope, emergencies, self-harm, vulnerable groups), and we make no coverage claim
beyond them. No adversarial red-teaming has been performed to date; the safety suite of the
production QA campaign (Section~\ref{sec:qa}) is the only systematic safety testing so far, and we
identify structured red-teaming as immediate future work.

\section{Evaluation}\label{sec:eval}

\subsection{Retrieval benchmark (released)}
We release a \textbf{50-question Thai retrieval golden set}:\footnote{Release bundle (golden
set, production QA logs, frontier-probe logs, evaluation harness; data CC~BY~4.0, code
Apache-2.0): \url{https://huggingface.co/datasets/iapp/tswap-wellness-benchmark} (code mirror:
\url{https://github.com/kobkrit/tswap-wellness-benchmark}).} 50 Thai questions grounded to gold
document identifiers across the live collections (built 2026-07-15), scored by Recall@$K$ with an
accompanying harness, alongside a sample of 50 realistic user questions. On this set the deployed
retriever attains \textbf{Recall@5 $= 0.88$} (44/50), with per-collection weak spots on
\texttt{thai\_herbs} and \texttt{packages} ($\approx 0.75$)---indicating that the herbal collection,
the paper's domain of interest, is also where retrieval most needs improvement. The repository also
contains a before/after harness for the metadata/range-filter change (a constraint-violation@$k$
metric), a regression gate, and a geospatial smoke suite.

\subsection{Deployment evaluation}
Four-week user acceptance testing ($n=120$: foreign and Thai tourists, wellness providers, tourism
operators) reported overall satisfaction $87.2\%$, user-assessed chatbot accuracy $86.5\%$, Thai
herbal-knowledge accuracy $90.2\%$, and provider-information trust $91.0\%$ (5-point Likert
instrument---user-assessed, not expert-graded; the objective correctness evidence is
Section~\ref{sec:qa}). Load testing sustained 200 concurrent users at $0\%$ error (platform-API
latency $\leq 527$\,ms avg; this excludes LLM generation, which the QA campaign measured at
14.1\,s median per answer), and a security assessment passed all 55 OWASP Top-10 (2025) cases.

\subsection{Production QA campaign (answer correctness)}\label{sec:qa}
Separately from the UAT, the platform team maintains a regression test list of \textbf{429 cases in
15 groups} covering the whole platform; the engine-correctness subset spans six groups: core
wellness content (ENG), colloquial-Thai geospatial provider search (LOC; 28 sub-categories such as
district/road/landmark/transit phrasing), multilingual behavior (LANG), conversational context
(CTX), safety and scope (SAFE), and input robustness (INP). Each case specifies an expected
behavior, and outcomes are graded \textsc{pass} / \textsc{odd} (anomalous) / \textsc{fail} against
it, executed manually against the production web client.

In a second-round campaign (August 2026), the \textbf{259} cases that passed round one were
re-executed: \textbf{236 (91.1\%)} reproduced \textsc{pass}, 21 became \textsc{odd}, and 2
\textsc{fail}---a direct test--retest consistency measurement that also exposes sampling
nondeterminism (the campaign log notes identical questions receiving different answers across
runs). Per group, the safety suite passed 19/21, multilingual 13/14, input robustness 11/11, core
wellness content 22/23, context 8/9, and geospatial search 163/181. A companion log tracks the
\textbf{56} problematic cases across both rounds (round 1: 46 \textsc{odd} / 7 \textsc{fail} / 3
\textsc{pass}; round 2: 13 \textsc{pass} / 40 \textsc{odd} / 3 \textsc{fail}); \textbf{47 of 56
(84\%)} fall in the colloquial geospatial group, with documented failure modes of false proximity
claims, unresolved landmarks and zones, and non-reproducible answers. End-to-end failures thus
concentrate exactly where retrieval is structurally hardest---colloquial place-name grounding---%
consistent with the retrieval-stage weak spots of the golden set.

\subsection{Frontier no-retrieval probe}\label{sec:probe}
Engine-side ablations (disabling retrieval or the safety prompt inside the deployed stack) require
serving-side switches; as a first controlled measurement we instead probed the engine's designated
alternative backend family---Gemini (2.5 Flash, Vertex AI)---\emph{without} retrieval, on
\textbf{71 real questions} drawn from the QA campaign (all 22 SAFE, all 23 ENG, 12 LOC, 14 LANG),
in two conditions: \emph{vanilla} (no system prompt) and \emph{+safety} (a reconstruction of the
TSWAP safety prompt of Section~\ref{sec:safety}). Outputs were scored by scripted surface checks
(hotline presence and position, referral/caution language, reply-language match) plus a manual
review of every SAFE and LOC answer by the authors; we label this a probe, not a benchmark, and
release the per-answer logs.\footnote{An initial run silently truncated answers because the
model's default hidden-reasoning budget consumed the output-token allowance---one more instance,
alongside the AWQ finding of Section~\ref{sec:eval}, of serving-side configuration materially
changing deployed behavior.}

\textbf{The safety prompt changes safety-critical behavior.} Vanilla, the frontier model answered
the dosage question (\tha{กินพาราได้ครั้งละกี่เม็ด}) with a complete paracetamol dosing schedule
including pediatric doses, and complied with both out-of-scope requests (writing Python code;
discussing politics): 3/22 clear violations of the expected safety behavior. With the safety
prompt, 0/22: the dosing request is refused and deferred to a pharmacist, and both out-of-scope
requests are declined with a redirect to wellness topics. On emergencies the difference is
\emph{prominence}: vanilla does eventually mention the Thai emergency number, but buries it after
a long differential-diagnosis-style explainer (1669 first appears at character 2{,}354 of a
3{,}596-character answer), whereas the +safety answer leads with it (character 141 of 280); the
self-harm hotline 1323 shows the same pattern (character 330 of 1{,}599 vs.\ 166 of 361). For a
user with chest pain, prominence is the metric that matters. Professional-referral language on
herbal content also rises from 14/23 to 22/23.

\textbf{Provider recommendation does not work without the knowledge base.} On the 12 colloquial
geospatial queries, the no-retrieval model produced \textbf{zero} verifiable provider
recommendations: 8/12 answers fell back to naming well-known national spa chains (Health Land,
Let's Relax, Oasis, Divana) with no address, hours, certification status, or confirmation that a
branch exists in the asked-about neighborhood or province, and 4/12 declined specifics entirely,
deferring the user to Google Maps or hotel staff. The deployed system answers the same query
family from 484 verified providers with geospatial filtering (163/181 \textsc{pass} in the QA
campaign). Multilingual reply behavior, by contrast, was unaffected by the prompt (13/14 in both
conditions), consistent with it being inherited from the base model in TSWAP as well.

Caveats: one frontier model, small $n$, author-graded surface metrics, and a reconstructed rather
than production prompt; the probe complements, not replaces, engine-side ablations.

\subsection{Serving findings}
Two deployment findings transfer beyond TSWAP. (1)~\emph{English-calibrated quantization degrades
Thai}: the community 4-bit AWQ checkpoint measurably corrupts Thai vowel and tone-mark generation,
motivating the post-generation quality guard (Section~\ref{sec:safety}) and a migration to FP8.
(2)~\emph{Forced retrieval is necessary}: with automatic tool selection the model silently skipped
retrieval on short queries; classifier-forced \texttt{tool\_choice} restored reliable grounding.

\subsection{Toward an answer-quality benchmark}\label{sec:proposed}
The released harness scaffolds the protocol we propose for a full Thai herbal answer-quality
benchmark: an expert-authored QA set tagged by topic and expected safety behavior, translated into
the eight languages with native back-translation \cite{xlingeval}; rubric-based grading
\cite{healthbench} by an LLM jury validated against native-speaker experts with chance-corrected
agreement; RAG faithfulness metrics per query-language \cite{ragas,medrag}; and
safe/problematic/unsafe labeling with red-teaming \cite{unsafe2025}. The engine's swappable
OpenAI-compatible backend makes the corresponding engine-side ablations cheap to run; they remain
future work.

\section{Discussion and Limitations}
TSWAP shows that a grounding stack around an \emph{unmodified} multilingual LLM can deliver a
deployed wellness advisor in a culturally specific, low-resource domain without fine-tuning.
Limitations, plainly: per-language quality is inherited from pretraining (a risk for the secondary
languages); the safety layer has no trained classifier or interaction checker; engine-side
ablations and per-language answer-quality results do not yet exist; retrieval is weakest exactly
on the herbal collection; end-to-end failures concentrate on colloquial geospatial queries and
answers are not fully reproducible across runs; and evaluation covers a single deployment. These
set the future-work agenda: the full answer-quality benchmark, engine-side ablations, FP8
migration, \texttt{thai\_herbs} retrieval improvements, and structured red-teaming.

\section{Ethics and Data Statement}
TSWAP forwards only data-minimized, wellness-relevant profile fields to the LLM (PDPA compliance).
It is positioned as general wellness guidance, not medical advice, and routes emergencies to Thai
hotlines (1669, 1323). Informed consent was obtained from all 120 UAT participants. The
50-question retrieval golden set, the production QA test logs, and the frontier-probe logs are
released with this paper; the herbal knowledge base itself is not part of the release.

\section{Conclusion}
TSWAP fills an empty niche in Thai NLP: no prior deployed, multilingual, retrieval-grounded Thai
traditional-medicine advisor existed, and no public Thai herbal benchmark. We release the first
such benchmark with production evaluation logs, and contribute grounding and serving findings we
expect to transfer to other tool-using RAG assistants in low-resource languages.

\section*{Acknowledgments}
This research project is financially supported by the Program Management Unit for Competitiveness
(PMUC) under grant number C05F680144.

\end{document}